\documentclass{article}

\RequirePackage[OT1]{fontenc}
\RequirePackage{amsmath}
\RequirePackage[numbers]{natbib}
\RequirePackage[colorlinks,citecolor=blue,urlcolor=blue]{hyperref}

\usepackage{amsmath}
\usepackage{mathtools}
\usepackage{amsthm}
\mathtoolsset{showonlyrefs}
\usepackage{amssymb}
\usepackage{calligra}
\usepackage[ruled,linesnumbered]{algorithm2e}
\SetKw{Break}{break}

\usepackage{amsfonts}

\title{How to gamble with non-stationary $\mathcal{X}$-armed bandits and have no regrets}
\author{Valeriy Avanesov\\
	WIAS Berlin\\
	\href{mailto:avanesov@wias-berlin.de}{avanesov@wias-berlin.de}
}

\date{\today}

\numberwithin{equation}{section}
\theoremstyle{plain}
\newtheorem{theorem}{}[section]

\newtheorem{lemma}{}[section]
\newtheorem{definition}{}[section]
\newtheorem{assumption}{}[section]
\newtheorem{remark}{}[section]

\usepackage[scr=boondox]{mathalfa}

\begin{document}
	\renewcommand{\thedefinition}{Definition \thesection.\arabic{definition}}
	\renewcommand{\thelemma}{Lemma \thesection.\arabic{lemma}}
	\renewcommand{\thetheorem}{Theorem \thesection.\arabic{theorem}}
	\renewcommand{\theassumption}{Assumption \thesection.\arabic{assumption}}
	\renewcommand{\theremark}{Remark \thesection.\arabic{remark}}
	\renewcommand{\thecorollary}{Corollary \thesection.\arabic{corollary}}

	\maketitle


\newcommand{\kdd}{k(\cdot, \cdot)}
\newcommand{\I}{\mathbb{I}}
\newcommand{\A}{A(\omega)}
\newcommand{\rhocp}{\rho^{\mathrm{CPD}}_n}
\newcommand{\rhoucb}{\rho^\mathrm{UCB}_t}
\newcommand{\inv}[1]{#1^{-1}}
\newcommand{\suffinv}{^{-1}}
\newcommand{\xstar}{X^*}
\newcommand{\ruleparam}{\xi}
\newcommand{\kstar}{k^*}
\newcommand{\X}{\mathcal{X}}
\newcommand{\F}{\mathcal{F}}
\newcommand{\N}{\mathbb{N}}
\newcommand{\R}{\mathbb{R}}
\newcommand{\eps}{\varepsilon}
\newcommand{\Ti}{T_i}
\newcommand{\K}{\mathcal{K}}
\newcommand{\Prob}[1]{\mathbb{P} \left\{#1\right\}}
\renewcommand{\v}{\mathrm{v}}
\newcommand{\x}{\mathrm{x}}
\renewcommand{\u}{\mathrm{u}}
\newcommand{\kk}{\mathrm{k}}

\newcommand{\RT}{R_T}

\newcommand{\fstar}{\mathscr{f}}
\newcommand{\fone}{\mathscr{f_1}}
\newcommand{\ftwo}{\mathscr{f_2}}
\newcommand{\fhat}{\mu}
\newcommand{\hatDelta}{\hat{\Delta}}
\newcommand{\hatDeltaSq}{\hat{\Delta}^2}
\newcommand{\brac}[1]{\left(#1\right)}
\newcommand{\cbrac}[1]{\left\{#1\right\}}
\newcommand{\True}{\mathtt{True}}
\newcommand{\False}{\mathtt{False}}
\newcommand{\g}{\mathfrak{g}}
\newcommand{\E}[1]{\mathbb{E}\left[#1\right]}
\newcommand{\Var}[1]{\mathbb{V}\mathrm{ar}\left[#1\right]}

\newcommand{\dotprod}[2]{\langle#1,#2\rangle}
\newcommand{\infnorm}[1]{\left\|#1\right\|_{\infty}}
\newcommand{\abs}[1]{\left|#1\right|}
\newcommand{\norm}[1]{\left\|#1\right\|}
\newcommand{\normSq}[1]{\left\|#1\right\|^2}
\newcommand{\knorm}[1]{\left\|#1\right\|_k}
\newcommand{\tildeDeltaSq}{\tilde{\Delta}^2}
\newcommand{\tildeDelta}{\tilde{\Delta}}
\newcommand{\Deltaf}{\Delta_f}
\newcommand{\lev}[1]{{#1}^{\le n}}
\newcommand{\prav}[1]{#1^{> n}}
\newcommand{\app}[1]{#1^{1..2n}}

\newcommand{\Ltwo}{\mathcal{L}_2}

\newcommand{\D}{\mathrm{D}}

\newcommand{\pluseq}{\mathtt{\mathrel{+}=}}
\newcommand{\plusplus}{\mathtt{++}}
\newcommand{\minuseq}{\mathrel{-}=}
\newcommand{\history}{\mathtt{history}}
\newcommand{\uniformlySampled}{\mathtt{uniformlySampled}}
\newcommand{\uniformlySampledn}{\mathtt{tail}}
\newcommand{\T}{\mathcal{T}}
\newcommand{\notT}{\mathcal{\bar{T}}}
\newcommand{\TT}{\mathbb{{T}}}
\newcommand{\Olog}[1]{\tilde{O}\left(#1\right)}
\newcommand{\LB}[1]{\Omega\left(#1\right)}
\newcommand{\tpower}{\frac{2\alpha + 3d(d+1)}{4\alpha+2d(d+1)}}
\newcommand{\betapower}{\frac{d(d+1)}{2\alpha + d(d+1)}}
	\begin{abstract}
		In $\X$-armed bandit problem an agent sequentially interacts with environment which yields a reward based on the vector input the agent provides. The agent's goal is to maximise the sum of these rewards across some number of time steps.
		The problem and its variations have been a subject of numerous studies, suggesting sub-linear and sometimes optimal strategies.
		The given paper introduces a new variation of the problem. We consider an environment, which can abruptly change its behaviour an unknown number of times.
		To that end we propose a novel strategy and prove it attains sub-linear cumulative regret.
		Moreover, the obtained regret bound matches the best known bound for GP-UCB for a stationary case, and approaches the minimax lower bound in case of highly smooth relation between an action and the corresponding reward.
		The theoretical result is supported by experimental study.
	\end{abstract}


\section{Introduction}
Numerous studies consider variations of an $\X$-armed bandit problem, a problem where an agent sequentially interacts with the environment, supplying a vector $X_t$ at each time step $t$ and receiving a reward $y_t$, depending (presumably) on $X_t$.
The reward is immediately made known to the agent, so the subsequent actions can be based on it.
Typically, the reward is obfuscated by some noise.

This model finds its use in applications such as clinical trials, pricing, finance, logistics, advertisement and recommendation \citep{Besbes,example1,example2,example3,example4,example5,example6} and has (along with stochastic optimization in general) unsurprisingly attracted immense attention in the recent decades.
Initially, though, the research was focused on {\it multi-armed bandits}, where only the actions $X_t$ from a finite set are feasible \citep{Gittins:89}.
Later researchers also turned to consideration of $\X$-armed bandits (with continuous feasible set $\X$) \citep{Srinivas2009,Bubeck:2011:XAB:1953048.2021053}.

Currently due to the ever-changing nature of our world, the studies on multi-armed bandits are increasingly more concerned with the environments changing their behavior over the course of time \citep{10.2307/2335176,gittens,whittle1981}.
For instance, an active line of work considers {\it restless} bandits, referring to a class of problems where the environment switches between the internal states according to a known stochastic law \citep{10.2307/3214163}. Also, there are settings where no such knowledge is available, like in the work by \cite{Besbes}, presuming a bound on the total variation of the mean reward associated with each arm, or implying no such bound, yet presuming these switches to be relatively rare \citep{Cao2018,pmlr-v99-auer19a}.
The latter is the setting we extend to the case of $\X$-armed bandits.
Particularly, we let the underlying relationship between the action $X_t$ and the reward $y_t$ to abruptly change.
The agent has no knowledge of when to expect such a change, neither any information on the nature of the change is revealed. As usual, in a bandit setting the goal is to minimize the {\it cumulative regret} -- a discrepancy between the received reward and the largest possible one. 

Formally, consider a number of stationary periods $K \in \N$ and a sequence of deterministic functions $\F \coloneqq \left\{\fstar_1, \fstar_2, ... , \fstar_K\right\}$ mapping from the convex compact $\X \subset \R^d$ to $\R$.
Further, we introduce a sequence of time points (being called {\it change-points}) $0=\tau_0<\tau_1 < \tau_2< ... < \tau_K=T$, when the environment switches between the functions $\fstar_i$, and denote the lengths of the stationary intervals as $T_i \coloneqq \tau_i - \tau_{i-1}$.
Also consider a piecewise constant map $\varkappa : \N \rightarrow \{1, 2, ...,K\}$, such that $\varkappa(1)=1$, $\varkappa(T) = K$, $\varkappa(i+1)=\varkappa(i)+1$ if $\exists j \in \N : i = \tau_{j}$ and $\varkappa(i+1) = \varkappa(i)$ otherwise.

At time points $t=\{1,2,...,T\}$ the agent consequently interacts with the environment, providing a vector $X_t\in\X$ and receives a reward
\begin{equation}
    y_t = \fstar_{\varkappa(t)}(X_t)+\varepsilon_t,
\end{equation}
where $\eps_t$ denotes i.i.d. centered noise.

Neither the functions $\fstar_i$ nor the change-points $\tau_i$ nor the number $K$ is known to the agent.
The goal is to minimize the {\it cumulative regret} $\E{R_T}$, where
\begin{equation}
    R_T \coloneqq \sum_{t=1}^T \max_{x \in \X} \left[\fstar_{\varkappa(t)}(x)\right] -  \fstar_{\varkappa(t)}(X_t).
\end{equation}
Here we emphasize the crucial difference between the problem we consider and the adversarial setting. 
We assess the performance of the agent in comparison to an oracle, choosing an optimal strategy for each stationary period, while the literature on adversarial environments competes with oracles whose choice of input remains the same during all $T$ steps, or allows for some number of switches at arbitrary time points \citep{Arora2012}.

The algorithm is said to be {\it no-regret}, if $\lim_{T \rightarrow +\infty} \E{R_T}/T = 0$, which is exactly what we achieve. Specifically, our contribution is outlined as follows
\begin{itemize}
  \item We propose a novel approach for a non-stationary $\X$-armed bandit problem.
  \item We establish a sub-linear bound on the cumulative regret under mild assumptions.
  \item The upper bound matches such for GP-UCB \citep[see][]{Chowdhury2017,Srinivas2009} for the stationary case, which implies the transition to the non-stationary setting came free of charge (asymptotically).  
  \item Our approach is adaptive to the number of change-points (unlike in \citealp{Cao2018}, where $K$ is deemed known). This adaptiveness is not achieved via keeping track of the number of change-points detected \citep[as in][]{pmlr-v99-auer19a}, yielding consistent behavior across all stationary intervals.  
  \item The approach is also adaptive to the extent of the change (formalized by \eqref{extentdef}), unlike the method suggested in \citep{Cao2018}.
  \item The algorithm is not adaptive to the horizon $T$, nevertheless, it is robust to misspecification of this parameter. Namely, the approach is driven by $\log T$, being more tolerant to an incorrect choice of horizon, than the approaches suggested by \cite{Cao2018, pmlr-v99-auer19a}, relying on the specification of $\sqrt{T}$.  
  \item The theoretical findings are verified empirically. A comparative study has also been conducted.
  \item As a byproduct, we propose an algorithm, generally suited for detection of a change-point in regression, not only in a bandit setting.
\end{itemize}
The paper is organized as follows. Section \ref{secstrat} introduces the suggested approach along with the necessary background and is followed by a rigorous theoretical study given in Section \ref{secth}. The theoretical results are put to a test in Section \ref{secexp} describing the empirical study. We conclude the paper with Section \ref{secfut} outlining the possible directions for future research.



\section{The proposed strategy}\label{secstrat}
This section presents the proposed algorithm in sub-section \ref{strategy}. 
We also develop a novel change-point detection algorithm as its necessary building block and describe it in sub-section \ref{seccp}. 
Both of these algorithms rely on Gaussian Process Regression, therefore we open the section with a brief description of this well-known technique.

\subsection{Background: Gaussian Process Regression}
In the given study we rely on a well known black-box non-parametric approach known as Gaussian Process Regression \citep{Rasmussen2006}.
Formally, we model the noise with a normal distribution and impose a zero-mean Gaussian Process prior with covariance function $k(\cdot, \cdot)$ on the regression function. Then for a sequence of covariates $X_1, X_2, \dots, X_n$ we have
\begin{equation}\label{key}
\begin{split}
f &\sim \mathcal{GP}\left(0,\sigma^2 \brac{n\rho}\suffinv k(\cdot, \cdot) \right), \\
y_j &\sim \mathcal{N}({f(X_j)},{\sigma^2}) \text{ for } j \in 1..n,
\end{split}
\end{equation}
where $n$ is the number of covariate-response pairs under consideration and $\rho$ is a regularization parameter.

For a given covariate $\xstar$ the predictive distribution is also Gaussian with mean
\begin{equation}
    \mu_* = \mu(\xstar)= \kstar \inv{\K} y
\end{equation}
and variance
\begin{equation}
    \sigma_*^2 = \sigma^2 \brac{n\rho}\suffinv \cbrac{k(\xstar, \xstar) - \dotprod{\kstar \inv{\K}}{\kstar}},
\end{equation}
where $y=[y_i]_{i=1..n}$, $\K=[k(X_i, X_j) + n\rho \delta_{ij}]_{i,j=1..n}$, $\kstar = [k(\xstar, X_i)]_{i=1..n}$ and $\delta_{ij}$ is the Kronecker symbol.

\subsection{Change-point detection procedure}\label{seccp}
Our approach requires a change-point detection procedure as its crucial building block. To that end we suggest Algorithm \ref{CPdetection}.
Given a sequence of covariate-response pairs $\{(X_t, y_t)\}_{t=1}^{2n}$, we train Gaussian Process Regression twice --- using the first and the second half of the given data respectively.
This way we obtain two predictive functions $\fhat_1$ and $\fhat_2$ and calculate the $\Ltwo$-distance between them
\begin{equation}\label{hatDeltaDef}
    \hatDelta^2 \coloneqq \int_{X\in \X} \brac{\fhat_1(X)-\fhat_2(X)}^2 dX.
\end{equation}
Finally, we compare the discrepancy against some predetermined threshold $\theta_n$.
Intuitively, if the covariate-response pairs were generated with the same functional relationship, $\hatDeltaSq$ should be small, while violation of this assumption should lead to larger values.

\begin{algorithm}[H]
	 \caption{CPD}\label{CPdetection}
    \KwData{Covariate-response pairs $\{(X_t, y_t)\}_{t=1}^{2n}$, threshold $\theta_n$, regularization parameter $\rhocp$}
    \KwResult{$\True$ if a change-point is detected, $\False$ otherwise}
    $\mu_1(\cdot) \leftarrow $ train GPR on $\{(X_t, y_t)\}_{t=1}^n$ with $\rhocp$

    $\mu_2(\cdot) \leftarrow $ train GPR on $\{(X_t, y_t)\}_{t=n+1}^{2n}$ with $\rhocp$

    $\hatDelta^2 \leftarrow \int\limits_{X \in \X} \brac{\mu_1(X)-\mu_2(X)}^2 d X$

    \Return{$\hatDelta^2 > \theta_n$}
\end{algorithm}

\subsection{GP-UCB-CPD algorithm}\label{strategy}
A well-known approach called GP-UCB was proven by \cite{Srinivas2009} to attain sub-linear regret in a stationary setting.
The main idea behind the algorithm is to train Gaussian Process Regression  at each time point $t$, using the history of rewards.
Denote the obtained predictive mean $\mu_t(\cdot)$ and predictive variance $\sigma^2_t(\cdot)$.
The next input vector is chosen using the optimistic rule
\begin{equation}\label{UCBRule}
    X_t=\arg\max_{X\in \X} \mu_t(X)+\sqrt{\beta_t}\sigma_t(X),
\end{equation}
obtaining an exploration-exploitation trade-off, where $\beta_t$ are hyperparameters.

In a non-stationary setting we cannot hope for good performance of GP-UCB anymore, as non-stationarity of the underlying distribution violates the assumptions of GPR consistency results.
To that end we suggest to use Algorithm \ref{CPdetection} in order to detect a change and abandon the history acquired so far. 
Unfortunately, we cannot use the history acquired by \eqref{UCBRule} for change-point detection, as the chosen vectors might concentrate in the vicinity of $\arg\max_{X\in\X}\fstar_i(X)$, which will be the case after some number of change-free iterations. 
First, we use an adaptive (with no access to the number of change-points $K$) rule to decide if the step should be dedicated to uniform exploration (see line \ref{conditionline}). 
If the rule suggests to explore, an arm $X_t$ is chosen uniformly (line \ref{uniformline}) and appended to $\uniformlySampled$ along with the reward $y_t$.
After each uniform exploration step we run Algorithm \ref{CPdetection} for each even-sized $\uniformlySampledn$ of $\uniformlySampled$ (line \ref{runtails}-\ref{runtailsEnd}). 
If a change-point is detected for any $\uniformlySampledn$, we abandon all the data we have accumulated so far (lines \ref{detectfirst}-\ref{detectlast}). 
Should the condition in the line \ref{conditionline} be false, the rule \eqref{UCBRule} is used instead. 
In both cases the reward is stored along with the input in $\history$.
\begin{remark}
    The idea behind the rule, choosing the number of the uniform exploration steps (line \ref{conditionline}) can be back-ported into an earlier method suggested by \cite{Cao2018} for multi-armed bandits, which relies on the knowledge of the number of change-points $K$. 
\end{remark}

\begin{algorithm}[H]
	\caption{GP-UCB-CPD}\label{myalgo}
    \KwData{Convex compact $\X$, thresholds $\cbrac{\theta_n}$, horizon $T$, sequence of positive real numbers $\{\beta_t\}$, real parameter $\ruleparam>0$, regularization parameters sequences $\cbrac{\rhoucb}$ and $\cbrac{\rhocp}$}

    $\history \leftarrow [~]$  {// empty list}

    $\uniformlySampled \leftarrow [~]$ {// empty list}

    \For{$t \in 1,2,...,T$}{
    
        \eIf{$\abs{\uniformlySampled} \le \ruleparam\sqrt{\abs{\history}}$}{ \label{conditionline}
            Play $X_t \sim U[\X]$ \label{uniformline}

            Receive $y_t \leftarrow \fstar_{\varkappa(t)}(X_t) + \eps_t$

            Append $(X_t, y_t)$ to $\uniformlySampled$ and to $\history$

            \For{$n \in 1,2, ..., \lfloor \abs{\uniformlySampled}/2\rfloor $}{\label{runtails}
            	$\uniformlySampledn \leftarrow \uniformlySampled[\mathtt{-2n}: ]$ // take the last $2n$ elements
            	
            	\If{$\mathrm{CPD(}\uniformlySampledn, \theta_n, \rhocp)$}{\label{detector}
            		$\uniformlySampled \leftarrow [~]$\label{detectfirst}
            		
            		$\history \leftarrow [~]$ \label{detectlast}
            		  
            		\Break
            	}
            }\label{runtailsEnd}
            
        }{
            $t' = \abs{\history}$

            $\mu_{t'}(\cdot)$,  $\sigma^2_{t'}(\cdot) \leftarrow$ train GPR on $\history$ with $\rho^{\mathrm{UCB}}_{t'}$

            Play $X_{t} \leftarrow \arg\max_{X\in \X} \mu_{t'}(X)+\sqrt{\beta_{t'}}\sigma_{t'}(X)$

            Receive $y_t \leftarrow \fstar_{\varkappa(t)}(X_t) + \eps_t$ \label{exploitaionline}
            
            Append $(X_t, y_t)$ to $\history$
        }
    }
{}
\end{algorithm}

\section{Theoretical analysis of GP-UCB-CPD}\label{secth}


First of all, we assume the noise $\eps_t$ to have light tails. Formally, we presume them to be \mbox{sub-Gaussian}.
\begin{definition}[Sub-Gaussianity]
We say, a centered random variable $x$ is sub-Gaussian with $\g^2$ if
\begin{equation}
\E{\exp(sx)} \le \exp\brac{\g^2s^2/2}, ~~\forall s\in\R.
\end{equation}
We say, a centered random vector $X$ is sub-Gaussian with $\g^2$ if for all unit vectors $u$ the product $\dotprod{u}{X}$ is sub-Gaussian with $\g^2$.
\end{definition}
In the following we focus on the Matérn covariance function
\begin{equation}
    k(X,X') \coloneqq 2^{1-\alpha} r^\alpha B_\alpha(r)/\Gamma(\alpha),
\end{equation}
where $r = \sqrt{2\alpha}\norm{X-X'}/l$, $\alpha > 1$ controls smoothness, $l$ is the lengthscale  and $B_\alpha$ denotes modified Bessel function of the second kind.

Clearly, in order to quantify the difficulty of change-point detection we have to introduce a measure of discrepancy between the functions $\fstar_i$. To that end we employ $\Ltwo$-norm:
\begin{equation}\label{extentdef}
    \Delta^2 \coloneqq \min_{i=1..K-1} \normSq{\fstar_i - \fstar_{i+1}}.
\end{equation}
In the theoretical part of the paper we use $\norm{\cdot}$ to denote the Euclidean norm of a vector and an $\Ltwo$-norm of a function, $\infnorm{\cdot}$ denotes the sup-norm, while $\knorm{\cdot}$ stands for the norm the reproducing kernel Hilbert space induced by $\kdd$ is endowed with.
\begin{theorem}\label{mainth}
    Let $\eps_t$ be sub-Gaussian with $\g^2$ and let there exist a positive $F$ such that $$\ \sup_{i=1..K} \knorm{\fstar_i} \le F.$$ 
    Choose some positive $\ruleparam$, positive $c$,
    \begin{equation}
        \sigma^2 = 6\g^2\log T,
    \end{equation}
    \begin{equation}
        \rhoucb = \sigma^{2}t\suffinv \text{, } \rhocp = c n^{-\frac{2\alpha+d}{2\alpha+d+1}},
    \end{equation}
    \begin{equation}
	    \theta_n = C\brac{F^2 +\g \log^2 T} n^{-\frac{2\alpha+d}{2\alpha+d+1}}, \text{ where $C$ depends only on $c$ and $\kdd$}
    \end{equation}   
    and
    \begin{equation}\label{defbeta}
        \beta_t = \D t^{\betapower}  \log^4 T, \text{ where $\D$ depends only on $F$ and $\kdd$.}
    \end{equation}
    Finally, let $T , K \rightarrow +\infty$, $\Delta \rightarrow 0$  and assume there is enough space between the change-points
    \begin{equation}\label{cpspacingass}
    	\brac{\frac{\log T}{\Delta}}^{2+\frac{2}{2\alpha+d}} \frac{\sqrt{\max\limits_{i=1..K}T_i}}{\min\limits_{i=1..K}T_i} = o(1).
    \end{equation}
    Then  
    \begin{equation}\label{regretbound}
    \begin{split}
        	\E{R_T} &= O\brac{ \Delta^{-\brac{2+\frac{2}{2\alpha+d}} } \sum_{i=1}^KT_i^{\tpower} \log^3 T} \\
        &=O\brac{\Delta^{-\brac{2+\frac{2}{2\alpha+d}} } K^{\frac{2\alpha - d(d+1)}{4\alpha + 2d(d+1)}} T^{\tpower} \log^3 T}.
    \end{split}
    \end{equation}
\end{theorem}
We defer the proof to Appendix \ref{proofsec}.
Let us compare the obtained bound against the known results. For the sake of clarity we will use the $\Olog{\cdot}$ notation omitting the polylog factors.
As demonstrated by \cite{Chowdhury2017,Srinivas2009}, in a stationary case GP-UCB accumulates the regret of at most $\Olog{T^{\tpower}}$. 
Now consider a non-stationary setting and assume, the change-point locations $\tau_1, \tau_2, ..., \tau_{K-1}$ have been made known to the agent. In such a case we can obviously bound the regret as $\Olog{\sum_iT_i^{\tpower}}$. 
This is exactly the bound we obtained for GP-UCB-CPD under fixed $\Delta$\footnote{As examination of our proof reveals, we can also allow $\Delta$ to approach $0$ at some polynomial rate, still matching the GP-UCB bound. The detail is omitted for brevity.}
in the realistic setting of unknown change-point locations.
Therefore we conclude, the change-point detection comes with no asymptotic overhead. 
Next, consider the lower bound $\LB{T^{\frac{\alpha+d}{2\alpha+d}}}$ obtained by \cite{Scarlett2017} for the stationary case. 
Clearly, for $K$ stationary periods the lower bound is $\LB{\sum_i T_i^{\frac{\alpha+d}{2\alpha+d}}}$. This indicates GP-UCB-CPD does not achieve minimax optimality, yet the obtained rate is considered  \citep[see][]{Calandriello2019} to be closely following the lower bound. 
Moreover, in case of highly smooth functions ($\alpha \gg d$) the method is nearly optimal. 

Further, in the work by \cite{Cao2018} the length of stationary periods is presumed to be at least $\sim\sqrt{T}$. Really, up to logarithmic terms under fixed $\Delta$ we only require
\begin{equation}
	\sqrt{\max\nolimits_iT_i} \ll \min\nolimits_iT_i.
\end{equation}

The suggested choice of parameters indicates the need for $T$ to be known in advance, however the parameters depend only on $\log T$. This implies robustness of the algorithm to misspecification of $T$, exceeding such of the approaches suggested by \cite{Cao2018,pmlr-v99-auer19a}, explicitly depending on $\sqrt{T}$.

\begin{remark}
	The statement of \ref{mainth} involves a number of parameters that should be tuned by a practitioner. The tuning of $\D$ seems to be unavoidable as we rely on GP-UCB. In Theorem 6 by \cite{Srinivas2009} the suggested choice of $\beta_t$ relies on the unknown RKHS norm of the reward function, the information gain $\gamma_t$, which is bounded up to an unknown multiplicative constant (see Theorem 5 therein). Usually, guess-and-doubling is employed to that end. The tuning of $C$ and $c$ is pretty straightforward in practice. For instance, we can run cross-validation on the uniformly sampled points to choose the optimal regularisation parameter $\rhoucb$ (effectively choosing $c$) and estimate the typical values of $\hat {\Delta}^2$ and choose $\theta_n$ (therefore $C$) to be large enough.  	
\end{remark}


\section{Experimental study}\label{secexp}
\begin{figure}
	\centering
  \includegraphics[width=0.6\linewidth]{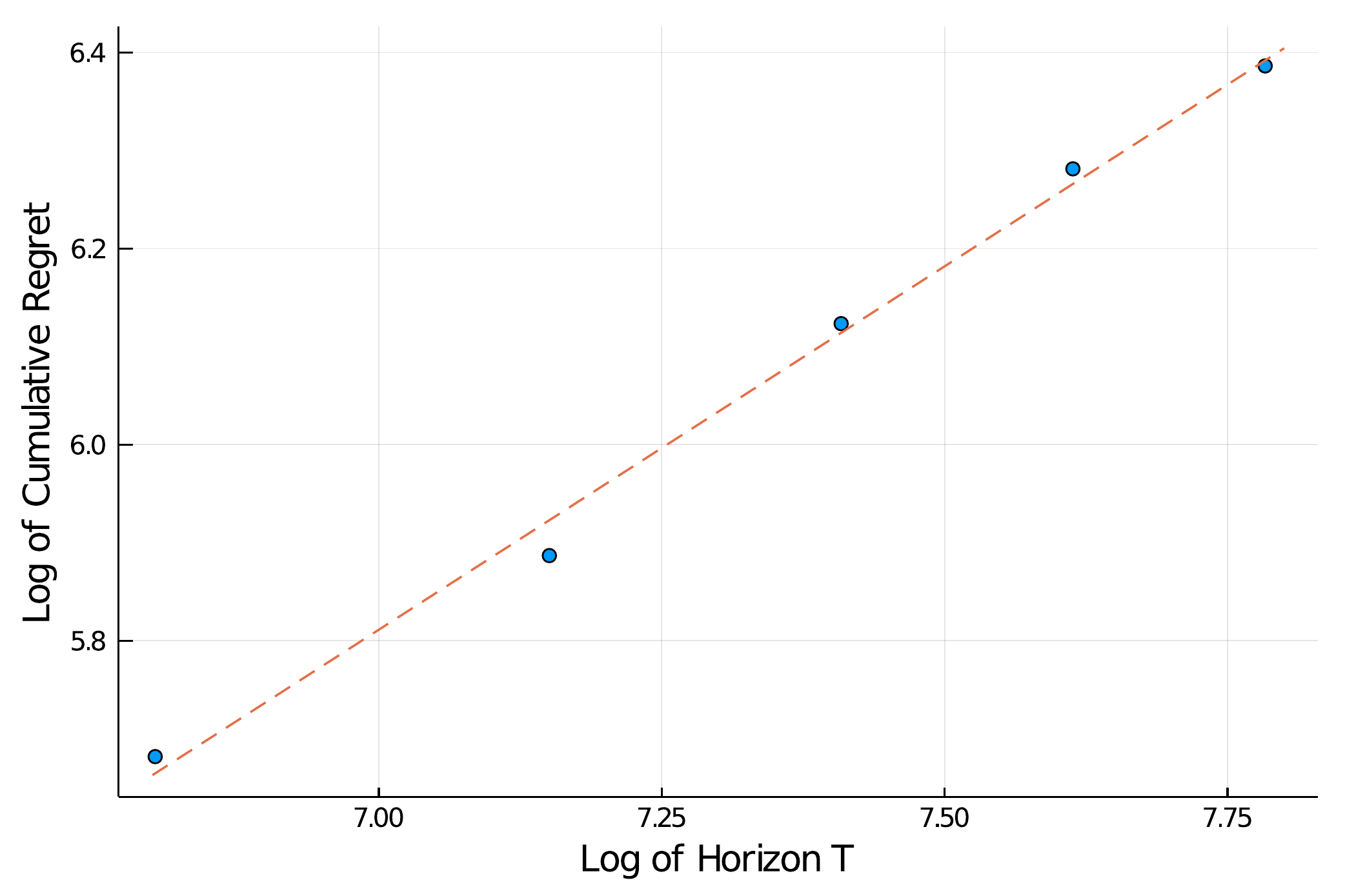}
  \caption{Here we present the dependence of cumulative regret $R_T$ on the horizon $T$ under fixed number of stationary periods $K=3$. The dashed line depicts the fitted curve $1.86T^{0.74}$. Both axes are in log scale.}
  \label{rangeTplot}
\end{figure}
\begin{figure}
	\centering
  \includegraphics[width=0.6\linewidth]{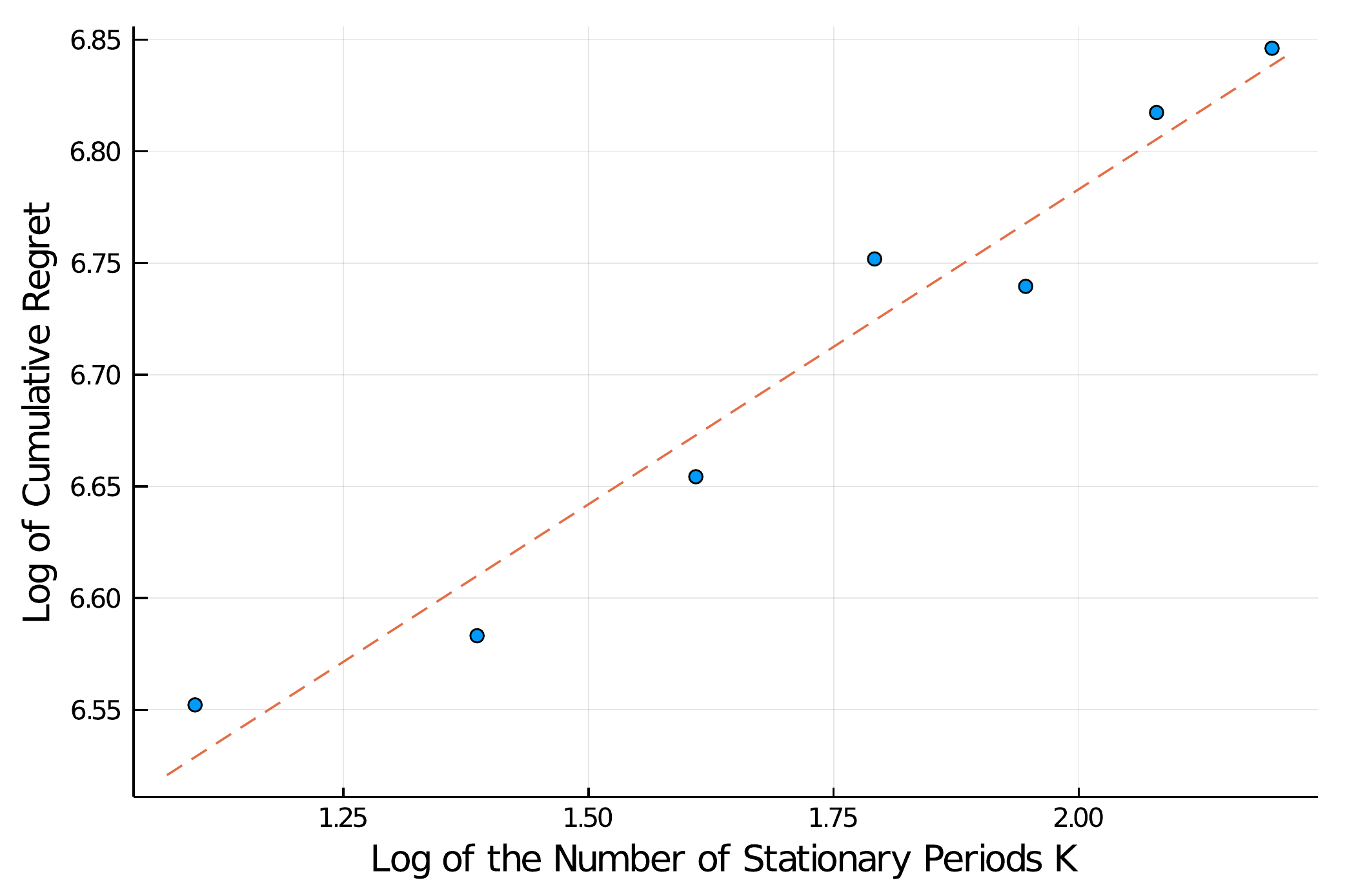}
  \caption{Cumulative regret $R_T$ for the number of stationary periods running from $3$ to $9$ under fixed horizon $T$. The fitted curve $502K^{0.282}$ is shown with a dashed line. Both axes are in log scale.}
  \label{rangeEpochPlot}
\end{figure}
\begin{figure}
	\centering
  \includegraphics[width=0.6\linewidth]{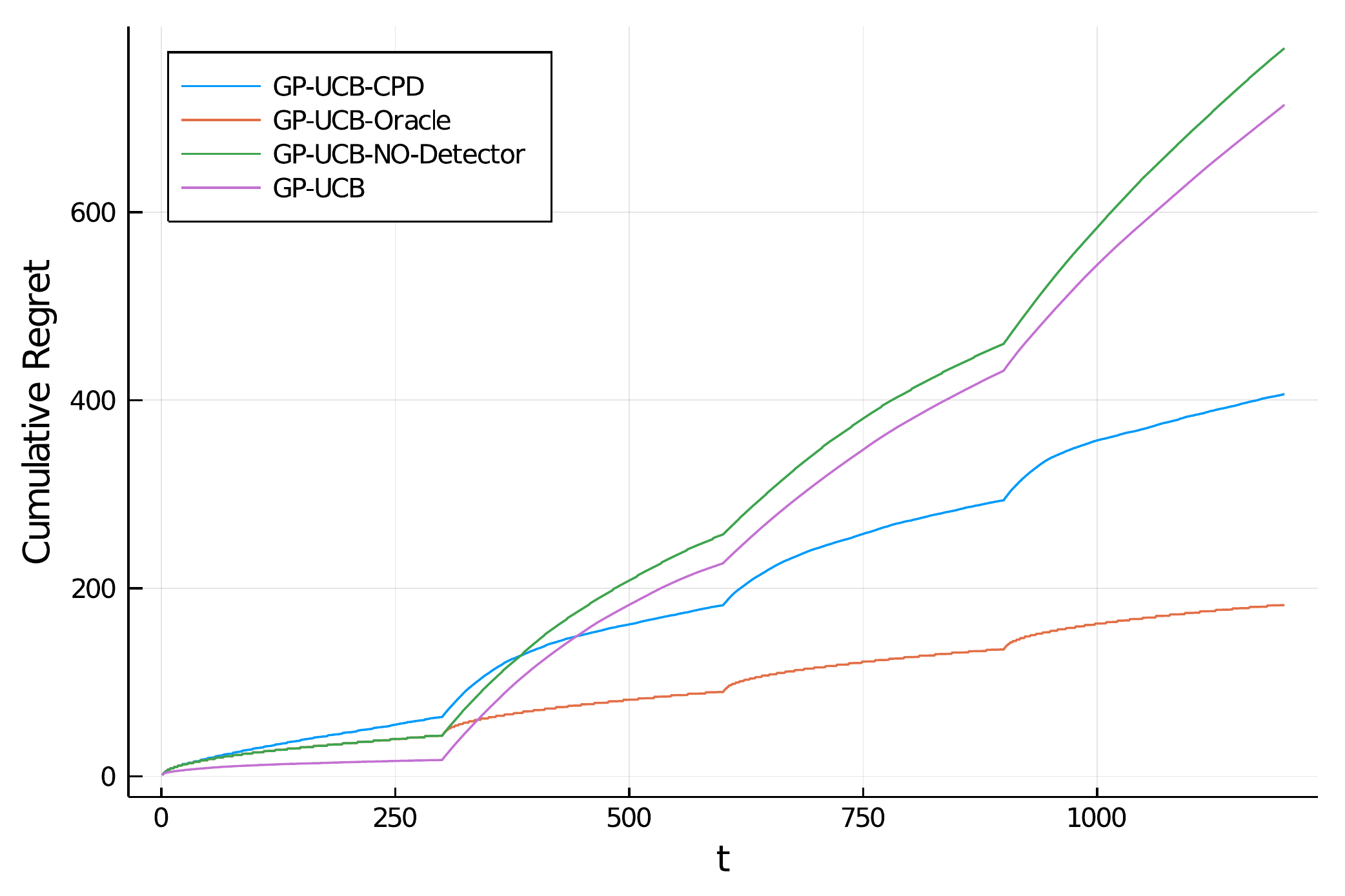}
  \caption{The plot demonstrates averaged cumulative regret of several algorithms interacting with an environment changing its behavior every $300$ points. 
  The compared algorithms are Algorithm \ref{myalgo} (denoted GP-UCB-CPD), Algorithm \ref{myalgo} using an oracle change-point detector (GP-UCB-Oracle), Algorithm \ref{myalgo} using a change-point detector, which never detects a change point (Algorithm \ref{myalgo} with $\theta_n=+\infty$, denoted GP-UCB-NO-Detector) and finally, GP-UCB, suggested by \cite{Srinivas2009} (Algorithm \ref{myalgo} with $\xi=0$).}
  \label{synthplot}
\end{figure}
In this section we support the theoretical results experimentally and present a comparative study. 
We consider a Matérn kernel with smoothness index $\alpha=5/2$ and lengthscale $l=1$, denoting it $k_0(\cdot, \cdot)$. The functions $\fstar_i$ are drawn independently from $\mathcal{GP}(0, k_0(\cdot, \cdot))$.  
The noise $\eps_t$ is independent, centered and Gaussian, its standard deviation is $0.05$. 
The chosen domain $\X=[0,5]$ is discretized into $1000$ evenly spaced points. 
The change-points are chosen to be evenly spaced, as this is obviously the most hostile setting maximizing the theoretical lower bound.
For all the experiments we choose the parameter controlling the portion of the steps, dedicated to the uniform sampling $\xi=\sqrt{3}$, the covariance function the Gaussian Process Regression uses is $k_0(\cdot, \cdot)$. 
$\D=0.02$, threshold of change-point detection algorithm $\theta_n =2.6 n^{-6/7}$.
The experiments are repeated $64$ times and the results are averaged.

In the first experiment we examine dependence of the cumulative regret $R_T$ on the horizon~$T$ under fixed  number of stationary periods $K$. 
Namely, we run Algorithm \ref{myalgo} for $K=3$ and $T \in \{900, 1275, 1650, 2025, 2400\}$. 
The results are shown in Figure \ref{rangeTplot}. 
We also fit a parametric curve $CT^{c}$ and the optimal power coefficient is $c=0.74$ with $95\%$ confidence interval being $(0.64, 0.84)$, demonstrating the dependence is clearly sub-linear.
In the experiment Algorithm \ref{myalgo} behaves in a strict accordance with Theorem \ref{mainth}, which suggests $c=0.786$.
 
In the next step we keep the horizon fixed $T=2700$ and assess $K=3,4,...,9$. 
 The results are reported in Figure \ref{rangeEpochPlot} along with the fitted curve $502K^{0.282}$. 
 The $95\%$ confidence interval for the power coefficient is $(0.21, 0.35)$. 
Again, the dependence is evidently sub-linear and strictly follows Theorem \ref{mainth}. 

We conclude the section with a comparative study. Here we choose $T=1200$ and $K=4$. 
The findings are presented in  Figure \ref{synthplot}. 
Algorithm \ref{myalgo} is denoted as GP-UCB-CPD. 
For the sake of comparison we also consider a version of Algorithm \ref{myalgo}, equipped with an oracle change-point detector and referred to as GP-UCB-Oracle. 
The other two algorithms we compare against are an algorithm equipped with a change-point detector, never detecting a change-point and the algorithm suggested by \cite{Srinivas2009}, abandoning the change-point detection (and the uniform sampling) altogether (via the choice $\ruleparam=0$). 
We call these approaches GP-UCB-NO-Detector and GP-UCB.

As we can see on the Figure \ref{synthplot}, before the first change-point GP-UCB performs best, which is not surprising as other algorithms have accumulated regret during uniform sampling. 
The fact that GP-UCB-Oracle and GP-UCB-CPD accumulate approximately equal regret during this period implies low probability of false positive decision by the change-point detection algorithm. 
After the first change-point GP-UCB-CPD performs notably worse, than GP-UCB-Oracle for some period of time, which is due to an unavoidable delay of change point detection.
This behavior is repeated when the subsequent change points happen.
In spite of the stellar score before the first change-point, performance of GP-UCB greatly deteriorates afterwards. 
Moreover, it turns out to perform only marginally better than GP-UCB-NO-Detector.

Overall, unsurprisingly the lowest average regret is achieved by GP-UCB-Oracle, the method aware of the location of the change-point. The second best is GP-UCB-CPD. 
\section{Future work}\label{secfut}
In the study we have considered a realistic setting of an $\X$-armed bandit problem and suggested a strategy achieving sub-linear cumulative regret and near-optimality for highly smooth functions. 
This conclusion follows from both theoretical and empirical studies. 
Yet, many questions remain unanswered. The lines of our future work can be foreseen as follows
\begin{itemize}
  \item As long as our approach relies on Gaussian Process Regression, whose performance deteriorates in high dimension, the suggested methodology is only effective in a low-dimensional setting. High-dimensional $\X$-armed bandits have already attracted researchers' interest in the past \cite{Djolonga2013}, but the non-stationary setting is yet to be analysed.   
  \item Switching from GPR-based to tree-based approaches (see \cite{Bubeck:2011:XAB:1953048.2021053}) can yield an approach attaining nearly-optimal performance for wider classes of functions $\fstar_i$.
  \item A different setting can also be considered (akin to the one suggested by \cite{Besbes}), where we allow the environment to change its behavior at every step $t$, yet impose a bound on the total variation.
  \item Gaussian Process Regression is notorious for its cubic time complexity, which renders it ineffective on large samples of data which are common nowadays. Thankfully, numerous linear-time approximate approaches have been developed (see \cite{Rasmussen2006}) and  can be used to alleviate the issue. Moreover, as we have to deal with ever-growing data sets, suggesting a distributed approach is another worthy step.
\end{itemize}
\section*{Acknowledgements}
The research of ``Project Approximative Bayesian inference and model selection for stochastic differential equations (SDEs)'' has been partially funded by Deutsche Forschungsgemeinschaft (DFG) through grant CRC 1294 ``Data Assimilation'', ``Project Approximative Bayesian inference and model selection for stochastic differential equations (SDEs)''.

Further, we would like to thank Vladimir Spokoiny, Alexandra Carpentier and Manfred Opper for the discussions which have greatly improved the manuscript.

	\appendix
\section{Proof of Theorem \ref{mainth}}\label{proofsec}
\begin{lemma}\label{mainlemma}
    Let $\eps_t$ be sub-Gaussian with $\g^2$,  denote 
    $$F \coloneqq \max_{i=1..K} \knorm{\fstar_i} $$ 
    and  choose 
    \begin{equation}
        \sigma^2 = 6\g^2\log T,
    \end{equation}
    \begin{equation}
        \rhoucb = \sigma^2 t^{-1},
    \end{equation}
    \begin{equation}\label{betaassbound}
            \beta_t = D  \brac{2 \knorm{\fstar}^2 + 300 \eta_t \log^3 \brac{tT^2}}, \text{ where } \eta_t \text{ is defined by \eqref{maxinfogain} and for any }   D\ge1 ,
    \end{equation}
  
    \begin{equation}
    	\rhocp = c n^{-\frac{2\alpha+d}{2\alpha+d+1}}, \text{ for any } c>0,
    \end{equation}
    
    \begin{equation}
    	\theta_n = C\brac{F^2 +\g^2 \log^2 T} n^{-\frac{2\alpha+d}{2\alpha+d+1}}, \text{ where } C \text{ depends only on $\kdd$ and $c$, }    
    \end{equation}     
    
    \begin{equation}\label{mindistancebound}
      n^*\brac{\frac{2}{\ruleparam}+\frac{1}{\ruleparam^2}}\sqrt{2\max\nolimits_iT_i} \le \min\nolimits_iT_i,
    \end{equation}
    where $n^*$ comes from Lemma \ref{cpbehlemma}.
     Then on a set of probability at least $1-6T^{-1}$
    \begin{equation}
        R_T \le {\sqrt{\frac{8D \log^3 T}{\log(1+\sigma^{-2})}} }\sum_{i=1}^KT_i^{\tpower} + \infnorm{k} F\brac{\ruleparam + 2n^*\brac{\frac{2}{\ruleparam}+\frac{1}{\ruleparam^2}}}\sum_{i=1}^K\sqrt{T_i}.
    \end{equation}
\end{lemma}
\begin{proof}
First of all, we bound single-step regret with
\begin{equation}\label{onestep}
	\sup_i{\infnorm{\fstar_i}} \le \infnorm{k} F.
\end{equation}
In order to estimate the delay of detection of $i$-th change point, consider an equation
\begin{equation}
	\sqrt{T_i + T_{i+1} + \gamma} - \sqrt{T_i+ T_{i+1}} = 1/\ruleparam,
\end{equation}
characterizing the maximum number of iterations $\gamma$ between two consecutive uniform sampling steps. Clearly,
\begin{equation}
	\gamma = \brac{2\sqrt{T_i+ T_{i+1}}+1/\ruleparam}/\ruleparam \le \brac{\frac{2}{\ruleparam}+\frac{1}{\ruleparam^2}}\sqrt{T_i+ T_{i+1}}.
\end{equation}
Hence, assumption \eqref{mindistancebound} ensures the agent can accumulate the necessary sample, meaning that assumption \eqref{ndeltaass} holds for each of the change-points. So, we can apply  Lemma \ref{cpbehlemma} to each instance of usage of Algorithm \ref{CPdetection}.
The statement of the lemma holds for all of them on a set of probability at least $1-4T^2e^{-\x}$. 
The rest of the argument is conditioned on this set.
Hence, the total delay of detection is at most
\begin{equation}
	\begin{split}
		\sum_i n^*\brac{\frac{2}{\ruleparam}+\frac{1}{\ruleparam^2}}\sqrt{T_i+ T_{i+1}} &\le \sum_i n^* \brac{\frac{2}{\ruleparam}+\frac{1}{\ruleparam^2}}\brac{\sqrt{T_i}+\sqrt{T_{i+1}}} \\
		&\le 2\sum_i n^*\brac{\frac{2}{\ruleparam}+\frac{1}{\ruleparam^2}}\sqrt{T_i}
	\end{split}
\end{equation}
and therefore due to \eqref{onestep} the regret accumulated in these periods is bounded with 
\begin{equation}\label{nondetectedbound}
	2\infnorm{k} F\sum_i n^*\brac{\frac{2}{\ruleparam}+\frac{1}{\ruleparam^2}}\sqrt{T_i}.
\end{equation}
Denote the regret accumulated between $\tau_i$ and $\tau_{i+1}$ in the line \ref{exploitaionline} (we exclude uniform sampling from consideration for now just like the iterations when the change has happened, but was not detected yet) as $R_i$. 
Now we apply Lemma \ref{boBreakBound} for each interval between the changes. 
Note, its claim holds with probability at least $1-2T^{-2}$, hence the claim holds for all the intervals simultaneously w.p. at least $1-2KT^{-2}$, but as long as $K\le T$, the probability is at least $1-2T^{-1}$.
\begin{equation}\label{boundForStableRegions}
\begin{split}
    \sum_{i = 1}^K R_i &\le \sum_{i = 1}^K \sqrt{8T_i\beta_{\Ti}\eta_{\Ti}/\log(1+\sigma^{-2})} \\
    &\le \sqrt{\frac{8\log^3 T}{\log(1+\sigma^{-2})}}\sum_{i=1}^KT_i^{\tpower} ,
\end{split}
\end{equation}
where we have also used Lemma \ref{th5} bounding $\eta_t$. 
Also we note, the regret of $\ruleparam \sum_i\sqrt{T_i}$ is accumulated during the uniform sampling. Incorporating this observation with \eqref{boundForStableRegions} and \eqref{nondetectedbound} and choosing $\x=3\log T$ we constitute the claim.
\end{proof}

\begin{proof}(of Theorem \ref{mainth})
	Clearly, for any choice $\ruleparam>0$ assumption \eqref{mindistancebound} holds for $T$ large enough due to assumption  \eqref{cpspacingass}.
	Due to Lemma \ref{th5} the choice of $\beta_t$ satisfies \eqref{betaassbound} for $\D$ large enough.
	Thus, Lemma~\ref{mainlemma} applies here and yields for some positive $D$, which depends only on $F$ and $\kdd$
	\begin{equation}
	\begin{split}
		R_T \le& \left\{ \sqrt{\frac{D \log^3 T}{\log(1+\sigma^{-2})}} \right.  \\
		& + \infnorm{k} F \left. \brac{\ruleparam +  \brac{\frac{C(F^2+\g^2\log^2 T)}{\Delta^2}}^ {1+\frac{1}{2\alpha+d}}\brac{\frac{4}{\ruleparam}+\frac{2}{\ruleparam^2}}} \right\}\sum_{i=1}^KT_i^{\tpower} 
	\end{split}
	\end{equation}
	on a set of probability at least $1-6T^{-1}$.
	Using the fact that 
	$$R_T \le T \sup_i{\infnorm{\fstar_i}} \le  T \infnorm{k} F$$ 
	we have
	\begin{equation}
	\begin{split}
		R_T \le& \left\{ \sqrt{\frac{D \log^3 T}{\log(1+\sigma^{-2})}} \right.  \\
		& + \infnorm{k} F \left. \brac{\ruleparam +  \brac{\frac{C(F^2+\g^2\log^2 T)}{\Delta^2}}^ {1+\frac{1}{2\alpha+d}}\brac{\frac{4}{\ruleparam}+\frac{2}{\ruleparam^2}}} \right\}\sum_{i=1}^KT_i^{\tpower} \\
		 + &6 \infnorm{k} F.
	\end{split}
	\end{equation}
	But 
	\begin{equation}
		{\frac{1}{\log(1+\sigma^{-2})}} = O\brac{\log T}
	\end{equation}
	and hence asymptotically we have
	\begin{equation}
		\begin{split}
			\E{R_T} &= O\brac{{  \brac{\log^{3}T + \brac{\frac{\log T}{\Delta}}^ {2+\frac{2}{2\alpha+d}}}  }\sum_{i=1}^KT_i^{\tpower}} .
		\end{split}
	\end{equation}
	A trivial observation $2+\frac{2}{2\alpha+d} < 3$ gives the first line of the claim.
	Optimization over $T_i$ under the constraint $\sum_i T_i = T$ yields the second line of the claim. 
\end{proof}

\section{Analysis of UCB rule}
This section adapts the regret bound for GP-UCB obtained by \cite{Srinivas2009}. In this section we assume the environment is stationary, i. e. for $t = 1,2,...,T$
\begin{equation}
    y_t = \fstar(X_t) + \eps_t.
\end{equation}
Denote the set of time-steps when condition in the line \ref{conditionline} of Algorithm \ref{myalgo} computes to $\True$ as $\T$, its complement as $\notT$ and $\TT \coloneqq \{1,2,...,T\}$.
Further, for a sequence of real values $\{a_i\}$ and a set of indices $E$ we write $a_E \coloneqq [a_i]_{i\in E}$.

Throughout this section whenever GPR is employed, the regularisation parameter $\rho$ is chosen as $\rho = \rhoucb \coloneqq \sigma^2 t\suffinv$ given a training sample of size $t$. So, effectively a prior $f\sim \mathcal{GP}(0, k(\cdot,\cdot))$ is imposed.

Here we employ the concept of information gain, defined as mutual information between the function $f\sim \mathcal{GP}(0, k(\cdot,\cdot))$ and the observations $y_\notT$
\begin{equation}
    \I(y_\notT;f) \coloneqq H(y_\notT) - H(y_\notT|f),
\end{equation}
where $H(\cdot)$ denotes entropy and in our case
\begin{equation}
    \I(y_\notT;f) = \frac{1}{2}\log \det \brac{I_{\abs{\notT}}+\sigma^{-2}\K_{\notT}},
\end{equation}
where $\K_{\notT} = [k(X_t,X_{t'})]_{t,t' \in \notT}$.
In order to extend the results by \cite{Srinivas2009} for the case allowing for uniform sampling (see line \ref{uniformline}) we prove the following trivial lemma.
\begin{lemma}\label{triviallemma}
    \begin{equation}
        \I(y_\notT;f) \le \I(y_\TT;f).
    \end{equation}
\end{lemma}
\begin{proof}
    The claim follows from the fact that the eigenvalues of $I_{\abs{\notT}}+\sigma^{-2}\K_{\notT}$ and $I_{\abs{\TT}}+\sigma^{-2}\K_{\TT}$ are larger or equal to $1$, while $\abs{\notT} \le \abs{\TT}$.
\end{proof}
The next result connects the information gain and the predictive variance of GPR.
\begin{lemma}[Lemma 5.3 by \citealp{Srinivas2009}]\label{lemma53}
    \begin{equation}
        \I(y_{\notT}; f) = \frac{1}{2} \sum_{t \in \notT} \brac{1+\sigma^{-2}\sigma^2_t(X_t)}.
    \end{equation}
\end{lemma}
Now we are ready to assess the properties of the highest information gain possible from $n$ observations
\begin{equation}\label{maxinfogain}
    \eta_n \coloneqq \sup_{\{X_i\}_{i=1}^n} \I(y_{1..n} ; f).
\end{equation}
\begin{lemma}[extension of Lemma 7.1 by \citealp{Srinivas2009}]\label{lemma71}
    For an arbitrary positive $\zeta$
    \begin{equation}
        \frac{1}{2} \sum_{t \in \notT} \max \{\sigma^{-2}\sigma^2_t(X_t), \zeta\} \le \frac{2\zeta}{\log (1+\zeta)}\eta_{T}.
    \end{equation}
\end{lemma}
\begin{proof}
    The proof consists in combining Lemma \ref{lemma53}, Lemma \ref{triviallemma} and the fact that $\min\{r,\zeta\} \le \zeta\log(1+r)/\log(1+\zeta)$ for all positive $r$.
\end{proof}

Now we can replace Lemma 7.1 in the proof of Theorem 6 in \citep{Srinivas2009} with Lemma~\ref{lemma71}.
\begin{lemma}[Extension of Theorem 6 by \citealp{Srinivas2009}]\label{theorem6}
    Let $\delta \in (0,1)$, $\sup_{t \in \TT} \abs{\eps_t} \le \sigma$ and choose
    \begin{equation}
        \beta_t \ge 2\knorm{\fstar}^2 + 300 \eta_t\log^3 (t/\delta).
    \end{equation}
     Then on a set of probability at least $1-\delta$  for all $x\in\X$ for all $t$ for the predictive mean $\mu_t(\cdot)$ obtained based on $t$ observations 
    \begin{equation}
        \abs{\mu_t(x) - \fstar(x)} \le \beta_t^{1/2}\sigma_t(x).
    \end{equation}
\end{lemma}
Next, we extend the result for the case of sub-Gaussian noise.
\begin{lemma}\label{thoeorem6extended}
    Let $\eps_t$ be sub-Gaussian with $\g^2$, $\delta\in(0,1)$ and $\u>0$. Choose $$\sigma = \g\sqrt{2\brac{\u+\log T}}$$
    and
    \begin{equation}
        \beta_t \ge 2\knorm{\fstar}^2 + 300 \eta_t\log^3 (t/\delta).
    \end{equation}
    Then on a set of probability at least $1-\delta-\exp(\u)$  for all $x\in\X$ for all $t$ for the predictive mean $\mu_t(\cdot)$ obtained based on $t$ observations 
   \begin{equation}
       \abs{\mu_t(x) - \fstar(x)} \le \beta_t^{1/2}\sigma_t(x).
   \end{equation}
\end{lemma}
\begin{proof}
    Due to sub-Gaussianity for all $t$ for any positive $\x$
    \begin{equation}
        \Prob{\abs{\eps_t} > \x} \le 2 \exp \brac{-\frac{\x^2}{2\g^2}}
    \end{equation}
    and uniformly
    \begin{equation}
        \Prob{\sup_{t\le T}\abs{\eps_t} > \x} \le 2 T \exp \brac{-\frac{\x^2}{2\g^2}}.
    \end{equation}
    Change of variables yields for any positive $\u$
    \begin{equation}
        \Prob{\sup_{t\le T}\abs{\eps_t} > \g\sqrt{2\brac{\u+\log T}}} \le \exp(-\u).
    \end{equation}
    Finally, choose $\sigma = \g\sqrt{2\brac{\u+\log T}}$ and apply Lemma \ref{theorem6}.
\end{proof}

Using Lemma \ref{thoeorem6extended} instead of Theorem 6 by \cite{Srinivas2009} we extend Theorem 3 by \cite{Srinivas2009} in the desired way, bounding the regret of GP-UCB-CPD in the absence of change-points.
\begin{lemma}\label{boBreakBound}
    Let $\eps_t$ be sub-Gaussian with $\g^2$, choose
    \begin{equation}
        \sigma^2 = 2\g^2(\u+\log T)
    \end{equation}
    and
    \begin{equation}
        \beta_t \ge 2\knorm{\fstar}^2 + 300 \eta_t\log^3 (t/\delta).
    \end{equation}
    Then on a set of probability at least $1-\delta-\exp(-\u)$
    \begin{equation}
        R_T \le \sqrt{8T\beta_T\eta_T/\log(1+\sigma^{-2})}.
    \end{equation}
\end{lemma}

In conclusion, we cite a result bounding the information gain.
\begin{lemma}[Theorem 5 by \citealp{Srinivas2009}]\label{th5}
	Let $k(\cdot, \cdot)$ be Matérn covariance function with smoothness index $\alpha$. Then
	\begin{equation}
		\eta_T = O\brac{T^{\frac{d(d+1)}{2\alpha+d(d+1)}}\log T}.
	\end{equation}
\end{lemma}

\section{Formal treatment of Algorithm \ref{CPdetection}}
In this section we establish two theoretical results regarding our change-point detection procedure. Namely, Lemma \ref{lemmacpundernull} provides an upper bound on $\hatDeltaSq$ in the absence of a change-point, while Lemma~\ref{lemmacpunderalternative} gives its lower bound. These two results combined induce a proper choice of the threshold $\theta_n$ and the necessary half-sample size $n^*$.

First, assume $\{X_t\}_{t=1}^{2n} \overset{\text{iid}}{\sim} U(\X)$ and let
\begin{equation}
    y_t = \fstar(X_t) + \eps_t,
\end{equation}
where $\eps_t$ denotes i.i.d. centered noise.
\begin{lemma}\label{lemmacpundernull}
    Let $\eps_i$ be sub-Gaussian with $\g$, $k(\cdot, \cdot)$ be a Matèrn kernel with smoothness index $\alpha$.
    Choose 
    \begin{equation}
    		\rhocp = c n^{-\frac{2\alpha+d}{2\alpha+d+1}}
    \end{equation}
    for any $c>0$. Then for all $\x>1.3$
    \begin{equation}
    	\hatDeltaSq \le C\brac{\knorm{\fstar}^2 +\x^2\g^2} n^{-\frac{2\alpha+d}{2\alpha+d+1}}
    \end{equation}  
    with probability at least $1-4e^{-\x}$ for some C, which depends only on $\kdd$ and $c$.
\end{lemma}
\begin{proof}
    The proof consists in applying Lemma \ref{gpconsistency} twice, yielding concentrations of $\fhat_1$ and $\fhat_2$ around $\fstar$ and a piece of straightforward algebra.
    \begin{equation}
    \begin{split}
        \hatDeltaSq &=  \normSq{\fhat_1 - \fhat_2} \\
        & =  \normSq{\fhat_1 - \fstar  + \fstar - \fhat_2}\\
        & \le  \normSq{\fhat_1 - \fstar} +\normSq{\fstar - \fhat_2}\\
        & \le 2 \Deltaf^2 ,
    \end{split}
    \end{equation}
    where $\Delta_f$ comes from Lemma \ref{gpconsistency}.
\end{proof}

On the other hand, let $\{X_t\}_{t=1}^{2n} \overset{\text{iid}}{\sim} U(\X)$ as before and let there be two functions $\fone$ and $\ftwo$ such that
\begin{equation}
    y_t = \fone(X_t) + \eps_t \text{ for } t \le n
\end{equation}
and
\begin{equation}
    y_t = \ftwo(X_t) + \eps_t \text{ for } t > n.
\end{equation}
Needless to say, the ability of the algorithm to detect a change-point depends on some measure of discrepancy between the two functions. We suggest to consider $\mathcal{L}_2$-norm
\begin{equation}
    \Delta^2 \coloneqq \normSq{\fone - \ftwo}.
\end{equation}

\begin{lemma}\label{lemmacpunderalternative}
	 Let $\eps_i$ be sub-Gaussian with $\g$, $k(\cdot, \cdot)$ be a Matèrn kernel with smoothness index $\alpha$.
	Choose 
	\begin{equation}
		\rhocp = c n^{-\frac{2\alpha+d}{2\alpha+d+1}}
	\end{equation}
    for any $c>0$. 
	Then for all $\x>1.3$
	\begin{equation}
		\hatDeltaSq \ge \Delta^2 - C\brac{\knorm{\fone}^2 + \knorm{\ftwo}^2 +\x^2\g^2} n^{-\frac{2\alpha+d}{2\alpha+d+1}},
	\end{equation}
    with probability at least $1-4e^{-\x}$ for some $C$, which depends only on $\kdd$ and $c$.
\end{lemma}
\begin{proof}
	Again, we will apply Lemma \ref{gpconsistency} twice and run some algebra
	\begin{equation}
	\begin{split}
		\hatDeltaSq & = \normSq{\fhat_1 - \fhat_2} \\
		& = \normSq{\fhat_1 - \fone + \brac{\fone - \ftwo} + \ftwo - \fhat_2} \\
		& \ge \normSq{\fone-\ftwo} - \Delta_{\fone}^2 -  \Delta_{\ftwo}^2,
	\end{split}
	\end{equation}
     where $ \Delta_{\fone}^2$ and $\Delta_{\ftwo}^2$ come from Lemma \ref{gpconsistency}.
\end{proof}

Finally, we are ready to describe the behavior of Algorithm \ref{CPdetection}.
\begin{lemma}\label{cpbehlemma}  Let $\eps_i$ be sub-Gaussian with $\g$, $k(\cdot, \cdot)$ be a Matèrn kernel with smoothness index $\alpha$ and denote
	$F \coloneqq \max\{ \knorm{\fstar}, \knorm{\fone}, \knorm{\ftwo}\} $.
	Choose 
	\begin{equation}
		\rhocp = c n^{-\frac{2\alpha+d}{2\alpha+d+1}}
	\end{equation}
    for any $c>0$ and
    \begin{equation}
        \theta_n = C\brac{F^2 +\x^2\g^2} n^{-\frac{2\alpha+d}{2\alpha+d+1}}
    \end{equation}    
    for any $\x> 1.3$ and for $C$, which depends only on $\kdd$ and $c$.
    Let  the  sample size $n$ be large enough
    \begin{equation}\label{ndeltaass}
         n \ge n^* \coloneqq 2 \brac{\frac{C(F^2+\x^2\g^2)}{\Delta^2}}^ {1+\frac{1}{2\alpha+d}}.
    \end{equation}
    Then no false alarm will be raised if the data is not subject to a change and if a change is present, detection is guaranteed with probability at least $1-4e^{-\x}$.
\end{lemma}
\begin{proof}
    Lemma \ref{lemmacpundernull} bounds $\hatDeltaSq$ in the absence of a change.
    Hence, the choice of the threshold allows for at most $4e^{-\x}$ first type error rate.
    Now under the alternative, using Lemma \ref{lemmacpunderalternative} we see that \eqref{ndeltaass} implies $\hatDeltaSq > \theta_n$ with probability at least $1-4e^{-\x}$.
\end{proof}

\section{Consistency of Gaussian Process Regression}
In this section we formulate a consistency result for Gaussian Process Regression. 
Consider a training sample of $n$ covariate-response pairs $(X_i, y_i)$, where $X_i \overset{iid}{\sim} U[\X]$ and 
\begin{equation}
	y_i = \fstar(X_i) + \eps_i
\end{equation}
for some centered i.i.d. noise $\eps_i$.
We will rely on the minimax-optimal bound obtained in \citep{pmlr-v125-avanesov20a}. 
It imposes the following two assumptions on the covariance function $k(\cdot, \cdot)$.
\begin{assumption}\label{keras1}
    Let there exist $C_\psi$  s.t. for eigenfunctions $\{\psi_j(\cdot)\}_{j=1}^\infty$ of covariance function $k(\cdot, \cdot)$
    \begin{equation}
       \max_j\infnorm{\psi_j} \le C_\psi.
    \end{equation}
\end{assumption}

\begin{assumption}\label{keras2}
Let for the eigenvalues $\{\lambda_j\}_{j=1}^\infty$ of covariance function $k(\cdot, \cdot)$ exist $b>1$, positive $c$ and $C$ s.t. $c j^{-b} \le \lambda_j \le C j^{-b}$ for all $j$.
\end{assumption}

\begin{lemma}[Theorem 1 in \citep{pmlr-v125-avanesov20a}]\label{gpconsistencyA} Assume $\eps_i$ are sub-Gaussian with $\g$, let Assumptions \ref{keras1} and \ref{keras2} hold. Choose 
	\begin{equation}
		\rho =  c n^{-\frac{b}{b+1}}
	\end{equation} 
for any $c>0$. Then  for all $\x>1.3$
\begin{equation}
	\norm{\fstar - \mu}^2 \le C\brac{\knorm{\fstar}^2 +\x^2\g^2} n^{-\frac{b}{b+1}}
\end{equation}  
with probability at least $1-2e^{-\x}$, where  $C$ depends only on $\kdd$ and $c$.
\end{lemma}
In the current study we are mainly concerned with the behavior of GPR estimator with Matérn kernel and under uniform distribution of covariates. In such case Assumptions \ref{keras1} and \ref{keras2} hold with $b=2\alpha+d$ (see \citep{Yang2017}). These observations propel the following corollary of Lemma \ref{gpconsistencyA}
\begin{lemma}\label{gpconsistency} Assume $\eps_i$ are sub-Gaussian with $\g$, let $k(\cdot, \cdot)$ be a Matérn kernel with smoothness index $\alpha$. Choose 
	\begin{equation}
		\rho = c n^{-\frac{2\alpha+d}{2\alpha+d+1}}
	\end{equation} 
    for any $c>0$. 
	Then for all $\x>1.3$
	\begin{equation}
		\norm{\fstar - \mu}^2 \le \Deltaf^2 \coloneqq C\brac{\knorm{\fstar}^2 +\x^2\g^2} n^{-\frac{2\alpha+d}{2\alpha+d+1}}
	\end{equation}  
	with probability at least $1-2e^{-\x}$ for some $C$, which depends only on $\kdd$ and $d$.
\end{lemma}

	\bibliographystyle{plain}
	\bibliography{main}

\end{document}